\documentclass[preprint,12pt]{elsarticle}

\usepackage{amssymb}
\usepackage{times}
\usepackage{soul}
\usepackage{amssymb}
\usepackage{amsthm}
\usepackage{url}
\usepackage[hidelinks]{hyperref}
\usepackage[utf8]{inputenc}
\usepackage[small]{caption}
\usepackage{graphicx}
\usepackage{booktabs}
\usepackage{algorithm}
\usepackage{algorithmic}
\usepackage{multirow}
\usepackage{comment}
\usepackage[figuresright]{rotating}
\usepackage{lineno}
\usepackage{amsmath,amsfonts}
\usepackage{array}
\usepackage[caption=false,font=normalsize,labelfont=sf,textfont=sf]{subfig}
\usepackage{textcomp}
\usepackage{stfloats}
\usepackage{url}
\usepackage{verbatim}
\usepackage{graphicx}
\usepackage{graphics}
\usepackage{times}
\usepackage{epsfig}
\usepackage{amsmath}
\usepackage{amssymb}
\usepackage{multirow}
\usepackage{booktabs}
\usepackage{color}
\usepackage{mathtools}

\journal{}

\begin{document}

\title{AdaTriplet-RA: Domain Matching via Adaptive Triplet and Reinforced Attention for Unsupervised Domain Adaptation}

\begin{frontmatter}
    
\author[label1]{Xinyao Shu}
\author[label2]{Shiyang Yan$^*$}
\author[label1]{Zhenyu Lu}
\author[label3]{Xinshao Wang}
\author[label4]{Yuan Xie}
\cortext[mycorrespondingauthor]{Corresponding author}
\address[label1]{School of Artificial Intelligence, 
Nanjing University of Information Science and Technology}
\address[label2]{Inria, Université Paris-Saclay}
\address[label3]{Zenith Ai}
\address[label4]{East China Normal University}

\begin{abstract}
Unsupervised domain adaption (UDA) is a transfer learning task where the data and annotations of the source domain are available but only have access to the unlabeled target data during training. Most previous methods try to minimise the domain gap by performing distribution alignment between the source and target domains, which has a notable limitation, i.e., operating at the domain level, but neglecting the sample-level differences. To mitigate this weakness, we propose to improve the unsupervised domain adaptation task with an inter-domain sample matching scheme. We apply the widely-used and robust Triplet loss to match the inter-domain samples. To reduce the catastrophic effect of the inaccurate pseudo-labels generated during training, we propose a novel uncertainty measurement method to select reliable pseudo-labels automatically and progressively refine them. We apply the advanced discrete relaxation Gumbel Softmax technique to realise an adaptive Topk scheme to fulfil the functionality. In addition, to enable the global ranking optimisation within one batch for the domain matching, the whole model is optimised via a novel reinforced attention mechanism with supervision from the policy gradient algorithm, using the Average Precision (AP) as the reward. Our model (termed \textbf{\textit{AdaTriplet-RA}}) achieves State-of-the-art results on several public benchmark datasets, and its effectiveness is validated via comprehensive ablation studies. Our method improves the accuracy of the baseline by 9.7\% (ResNet-101) and 6.2\% (ResNet-50) on the VisDa dataset and 4.22\% (ResNet-50) on the Domainnet dataset. {The source code is publicly available at \textit{https://github.com/shuxy0120/AdaTriplet-RA}}.
\end{abstract}


\begin{highlights}
\item To facilitate the domain matching in unsupervised domain adaptation, we propose an uncertainty-aware Triplet loss, with a novel uncertainty measurement method via a trainable adaptive Topk selection, to make a clearer decision for hard samples in the target domain. 

\item We propose a novel reinforced attention mechanism algorithm to enhance the feature representation and the inter-domain sample matching in the unsupervised domain adaptation task. 

\item Our proposed methods are validated via comprehensive experiments on several publicly available benchmark datasets with State-of-the-art results. 
\end{highlights}

\end{frontmatter}

\section{Introduction}
Deep learning methods have achieved tremendous success in computer vision applications, e.g., image classification~\cite{szegedy2015going,simonyan2014very,he2016deep,dosovitskiy2020image}, object detection~\cite{girshick2015fast,ren2015faster}, semantic segmentation~\cite{long2015fully,ronneberger2015u}. However, it is restricted to supervised learning where a well-labelled dataset is available. How to generalise well in a dataset with a different distribution remains a challenging problem. Recently, researchers have extensively studied the unsupervised domain adaptation (UDA)~\cite{venkateswara2017deep,tan2020class,wu2019domain}, where the labels are in source data but unavailable in the target data.

\begin{figure}
    \centering
    \includegraphics[width=0.9\linewidth]{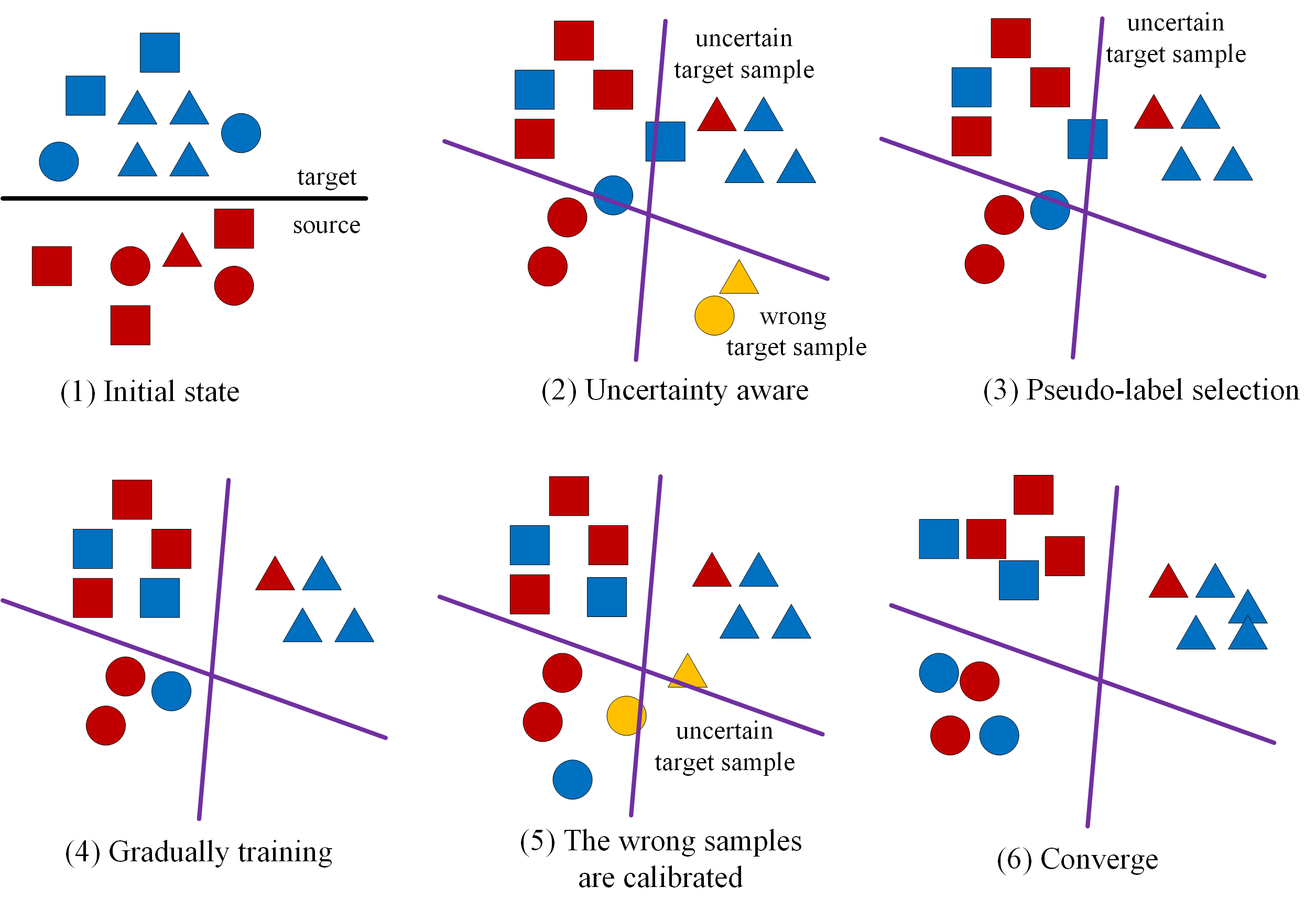}
    \caption{{An illustration of the idea of \textbf{\textit{AdaTriplet-RA}}: The different shape indicates different category. Initially, the source and target samples are separated via the domain boundary (1). When generating pseudo-labels on the target domain, there would be uncertain examples near the decision boundary and bad examples with extreme uncertainty (2). We then perform pseudo-label selection, ignoring the wrong samples (3). Through gradual training (4), the samples that are near the decision boundary are gradually becoming clearer, and the wrong samples are calibrated in the next step (5) until convergence (6).}}
    \label{fig:illustration}
\end{figure}

The challenges in unsupervised domain adaptation are twofold: it needs to ensure the feature representations are agnostic to domains; it also needs to keep the discriminating capability at the same time. Most of the previous research tries to {lower the empirical risk~\cite{vapnik1991principles} at the source domain} and perform the distribution alignment of the source and target domains, assuming lowering the actual risk in the target domain~\cite{ganin2015unsupervised,long2017deep}. For instance, adversarial learning for domain distribution alignment \cite{ganin2015unsupervised, tzeng2017adversarial, zhang2019domain} represents a classical line in this direction. Maximum Mean Discrepancy (MMD)-based methods \cite{yan2017mind, chen2019graph} is another commonly-applied approach for domain alignment. Unfortunately, on the one hand, the models are more accessible to over-fitted to the source task and do not generalise well to the target task; on the other hand, the alignment of sample space between domains is ignored to alleviate the domain shift. As a result, the source classifier can wrongly recognise target samples close to the decision boundary. Moreover, even with successful domain distribution alignment, the model neglects the sample-level similarity and has a higher possibility of negative transfer.


Hence we look directly at sample matching between domains, and specific questions can be raised: {can the inter-domain sample matching help the unsupervised domain adaptation task, and if so, how to realise it?} The answer to the first question is {Yes.} Previous image/text and image matching research \cite{li2019visual, lee2018stacked, zhang2018deep, chen2017multi, zhai2019defense} tries to exploit the sample matching for full-labelled data, in which the classification and retrieval performance are mutually promoted, as proved via many empirical results. This effectiveness of combining classification and matching in retrieval tasks demonstrates that sample-level matching: can both {promote recognition performance} and {perform modality/domain alignment} at the sample level. Both of them are what we want to achieve in unsupervised domain adaptation. Classical approaches \cite{li2019visual, lee2018stacked, zhang2018deep, chen2017multi, yuan2020defense} in retrieval tasks often utilise the metric learning loss as the principal matching loss. Metric learning tries to measure and manipulate the similarities between samples regardless of the domain/modality differences. 
Hence, the question remains: {How to realise domain matching?}
As in such an unsupervised learning setting, no labels are available for the target domain samples. Previous research either utilises constrained method \cite{deng2020rethinking} to solve this issue or class-wise matching with probability-based methods like MMD class-wise discriminant \cite{zhang2018unsupervised}. Alternatively, we directly apply Triplet loss for domain matching with the following contributions.

We propose an uncertainty-aware adaptive Triplet loss for the unsupervised domain adaptation. It successfully achieves the matching of semantically similar samples from different domains. We use the classifier trained on the source domain data to predict the pseudo-labels for target domain samples and apply the distance metric learning loss to optimise the model. One of the biggest challenges is that these pseudo-labels contain much noise, i.e., incorrect classification results, which introduce serious bias to the metric learning algorithm.

We introduce a robust {pseudo-label selection} method, where we define a new way of {uncertainty measurement} of the classification results in the target domain. We first define {prototypes} in the source domain, one prototype per class. Each prototype is computed via the average of all the features within one epoch whose performance is better than the average over the whole training stage. We select the reliable pseudo-labels and only use their corresponding samples in Triplet loss. Usually, we need to manually set a threshold value to avoid too many uncertain labels, which could cause a catastrophic effect on training. The selection of high certainty only targets pseudo-labels. The extremely uncertain labels cannot be relied on, as they are usually wrong. Hyper-parameter tuning for the threshold is laborious and yields poor performance. In this paper, we propose a novel trainable Topk scheme in which the threshold is determined via a Gumbel Softmax~\cite{jang2016categorical} discrete relaxation technique. Gumbel Softmax is a continuous distribution that can be smoothly annealed into a categorical distribution whose parameter gradients can be easily computed via the reparameterisation trick~\cite{jang2016categorical}. By doing this, the trainable Topk can automatically select the desired reliable pseudo-labels. We apply the selected pseudo-labels in the Triplet loss and their corresponding uncertainty value as the margin. The higher the uncertainty, the higher the margin should be. The training process is illustrated in Figure~\ref{fig:illustration}: {Initially, the data is only separated via domain boundaries. As the training is performed, the classifier tries to decide on the target data, but with uncertain samples and extremely uncertain samples with wrong pseudo-labels. Our model can gradually calibrate the wrong samples through the pseudo-label selection scheme and match the uncertain samples with the adaptive Triplet loss. Upon convergence, the source and the target samples are successfully classified via the category classifier.}

{Meanwhile, Triplet loss~\cite{schroff2015facenet} is a kind of batch-based optimisation goal for a certain model, which does not or in-comprehensive consider the global ranking quality of the whole batch. In many image matching or retrieval tasks, the performance evaluation is based on {Average Precision (AP)}, which comprehensively evaluates the global ranking performance. Nevertheless, AP is non-differentiable because of the discreteness, and the non-convexity \cite{chen2019towards}. In other words, AP cannot easily be approximated by discrete relaxing techniques such as straight-through estimator~\cite{cheng2019straight}. We want to optimise for a higher AP value during the training, whose solution lies in reinforcement learning (RL). Meanwhile, previous attention mechanism~\cite{xu2015show,anderson2018bottom} often treat attention weights as neurons in the network, which lacks strong supervision. Our approach tries to blend attention supervision with reinforcement learning. The attention weights generation is modelled as a Markov Decision Process (MDP)~\cite{sutton2018reinforcement}, and optimised via a simple policy gradient (PG) algorithm~\cite{sutton2018reinforcement, sutton1999policy}. We treat the AP as the reward in the PG algorithm, naturally solving AP's non-differentiable and non-convexity problems in usual supervised learning. Similar to the proposed adaptive distance metric learning, we adjust the reward with the certainty value to compensate for the noise in the pseudo-labels. The higher the certainty value, the higher reward should be given to the model. Note that the certainty is Top selected, and the AP is instance-level, with each sample having one AP result. Our model is termed as \textbf{\textit{AdaTriplet-RA}}, meaning Adaptive Triplet loss and Reinforced Attention.}

To summarise, the contributions of our paper are threefold: 
{
\begin{itemize}
    \item To facilitate the domain matching for the unsupervised domain adaptation task, we propose an uncertainty-aware Triplet loss to refine the pseudo-labels progressively. The scheme has a novel uncertainty measurement method realised via a trainable adaptive Topk selection to make a clearer decision for hard samples in the target domain.
    \item We propose a novel reinforced attention mechanism algorithm to enhance the feature representation and domain matching. The reinforced attention uses the Average Precision (AP) as the reward, which is also adaptively adjusted with uncertainty values. Reinforced attention plays a critical role in domain matching and improves performance.
    \item The proposed method ``AdaTriplet-RA" significantly improves the baseline methods and validates that successful domain matching can indeed boost the unsupervised domain adaptation task. 
\end{itemize}
}


\section{Related Work}
\subsection{Unsupervised Domain Adaptation}
\paragraph{General Methods}
Unsupervised domain adaptation (UDA) transfers knowledge from a labelled source domain to an unlabeled target domain. Existing unsupervised domain adaptation methods focus on image classification. The mainstream approaches tend to address unsupervised domain adaptation by learning domain-invariant representation, to which our method belongs. There are mainly two kinds of approaches to learning domain-invariant features. \cite{tzeng2014deep,long2017deep,long2015learning,yan2017mind,weighted2020tmm,su2021tmm,ding2022tmm} measure the domain similarity via Maximum Mean Discrepancy (MMD)~\cite{borgwardt2006integrating}. Another line of research learns domain-invariant features using neural model-based learning, e.g., adversarial training. A representative work is the DANN~\cite{ganin2015unsupervised}. This approach applies an implicit adversarial training scheme to learn domain-invariant representation via a gradient reversal layer and a discriminator. Subsequently, research follows this direction and yields good performance in unsupervised domain adaptation task~\cite{cui2020gradually, chen2020adversarial,long2017conditional,tzeng2017adversarial, sankaranarayanan2018generate}. Notably, SymNets~\cite{zhang2019domain} proposes a symmetric object classifier that plays the role of domain discriminator. Alternatively, Zhang et al. \cite{zhang2018unsupervised} directly targeting at class-wise matching by minimising MMD-based class-wise fisher discriminant across domains. Zhao et al. \cite{zhao2022source} address the noisy pseudo-labels for source-free DA tasks where only the pre-trained model and the target data are available during training. MMAN~\cite{MMAN} introduces semantic multi-modality representations learning into adversarial domain adaptation and uses multi-channel constraints to capture fine-grained knowledge categories. Deng et al.~\cite{deng2022tmm} uses disentanglement for adversarial domain adaptation to extract more transferable high-level semantic features.
Our research builds on the baseline of SymNets \cite{zhang2019domain}. However, it focuses on the improving impact of the domain matching, also addressing the noisy pseudo-label problem \cite{zhao2022source}, with a significant improvement on unsupervised domain adaptation. 

\paragraph{Domain Matching in Domain Adaptation}
There is research analysing the feasibility, problem and performance of the domain matching in domain adaptation tasks~\cite{deng2020rethinking, xie2018learning, laradji2020m, wang2022cross, sharma2021instance}. In particular, Deng et al.~\cite{deng2020rethinking} propose a similarity-guided constraint (SGC) for domain matching via the Triplet loss and emphasis the importance of domain matching. Xie et al.~\cite{xie2018learning} utilise semantic loss and adversarial domain matching for the unsupervised domain adaptation task. They propose to apply the running average for the centroid formulation and conduct centroid alignment with squared Euclidean distance loss. M-ADDA~\cite{laradji2020m} performs a different approach for sample-level matching in domain adaptation: adjusting the margin in metric learning loss with uncertainty. Wang et al.~\cite{wang2022cross} seek an alternative solution from the Contrastive loss~\cite{wu2018unsupervised} and formulate pseudo labels for the target domain, improving the existing approaches. Xu et al. \cite{xu2019unsupervised} apply importance sampling for both the domain and class-level matching in unsupervised domain adaptation. Sharma et al.~\cite{sharma2021instance} propose an instance matching scheme for domain adaptation, utilising not only the multi-sample contrastive loss but also cross-entropy. Li et al. \cite{li2021adadc} targets at unsupervised domain adaptation person re-ID task. They propose to address noisy labels and progressively refine them in deep clustering. Meng et al. \cite{meng2022exploring} exploit the label structural information via iterative clustering and pseudo labels for unsupervised domain adaptation. Our methods differ from this research in two perspectives: our uncertainty is selected with an adaptive threshold; we additionally apply reinforced attention with AP as the reward for better sample matching.

\subsection{Distance Metric Learning} 
Metric learning is a spatial mapping method which can learn a feature space. In this space, it makes the feature distance of similar samples smaller. Conversely, it makes the feature distance of different samples larger to distinguish them. 
Distance metric learning plays a significant role in a variety of computer vision applications, such as image retrieval~\cite{sohn2016improved}, cross-modal image-text matching~\cite{lee2018stacked}, person re-ID~\cite{hermans2017defense},  and transfer learning~\cite{oh2016deep}. Current research on distance metric learning focuses on the loss functions, e.g., Triplet loss~\cite{schroff2015facenet, hermans2017defense}, N-pair-mc~\cite{sohn2016improved}. There is also research work exploiting the mining techniques to consider the relationships between data samples, e.g., lifted structured~\cite{oh2016deep}, ranked list loss~\cite{wang2019ranked}. Among them, the Triplet loss is one of the most widely-used metric learning functions in varying tasks, given its simplicity and stability. However, most of the previous distance metric learning methods focus on developing the loss functions and mining techniques during the learning process. We focus on the pseudo-label-based Triplet loss with adaptive margin.

\subsection{Visual Attention Mechanism} 
The visual attention mechanism~\cite{xu2015show} has been widely applied in many computer vision applications. Notably, the bottom-up attention model~\cite{anderson2018bottom} is the current mainstream for image captioning, visual question answering, and image-text matching. However, there needs to be more research on supervised attention. Gan et al.~\cite{gan2017vqs} propose a supervised attention scheme for visual question answering using attention annotations. Kamigaito et al.~\cite{kamigaito2017supervised} also use attention annotations for supervised attention in natural language processing tasks. Instead, we propose a supervised attention mechanism based on reinforcement learning, which can optimise the attention module towards a specific goal such as AP. Also, the proposed attention module does not need any additional annotations.

\subsection{Discreteness Relaxation Techniques}
Usual neurons in deep learning models are continuous variables, which create a non-linear mapping between the inputs and outputs. There is also a family of stochastic discrete variables in neural networks \cite{xu2015show, yin2019understanding, chung2016hierarchical, jang2016categorical, kusner2016gans}.

 One has to rely on either reinforcement learning to realise sampling and exploration \cite{xu2015show, yan2021discrete}, or discreteness relaxation techniques like straight-through estimators \cite{bengio2013estimating, chung2016hierarchical}, and Gumbel techniques \cite{jang2016categorical, kusner2016gans} to train the discrete variables. Gumbel is a more efficient and effective technique in discreteness relaxation. 

Our model has two discrete variables: the optimising goal (AP) of reinforced attention and the $k$ for pseudo-label selection. Hence, we apply reinforced training and Gumbel techniques to solve these problems.

\section{The Proposed Method}
\begin{figure*}[h]
    \centering
    \includegraphics[width=\linewidth]{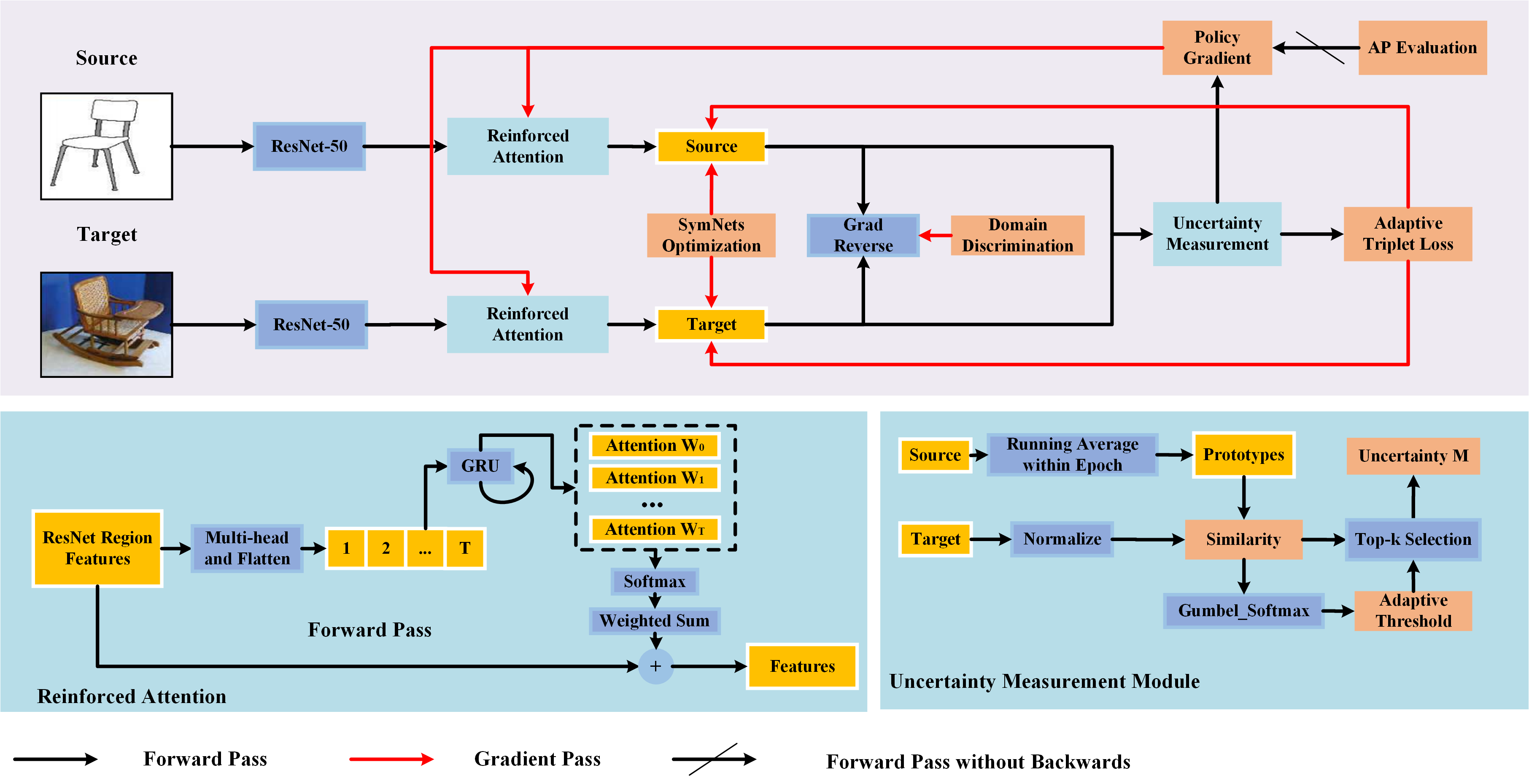}
    \caption{A schematic diagram of our model: The source and target images are fed into the ResNet-50 backbone network, followed by the extraction of the region. Based on the region features, we explore the reinforced attention via a GRU model and Softmax normalisation. The features are processed via the uncertainty measurement module to produce reliable pseudo-labels and corresponding uncertainty scores. The uncertainty score is also utilised in the policy gradient algorithm for reinforced attention optimisation and the adaptive Triplet loss. The whole model is optimised via multiple training losses.}
    \label{fig:system}
\end{figure*}

In this section, we introduce the proposed methods. We first briefly introduce the baseline model, then the regional representation, followed by a detailed illustration of the proposed adaptive triplet loss, the reinforced attention mechanism and the overall loss objectives. 

\subsection{Baseline Model.}
We apply the SymNets model~\cite{zhang2019domain,zhang2020unsupervised} as the baseline model for our research. The SymNets belong to the family of domain confusion methods~\cite{zhang2019domain}. 
SymNets is a symmetric network to overcome the limitation in the joint distribution of cross-domain aligned features and categories through two-level domain confusion loss. 
The category-level confusion loss improves over the domain-level one by driving the learning of intermediate network features to be invariant at the corresponding categories of the two domains. 
The design of a SymNet is based on a parallel task classifier $C^s$ and $C^t$. Assume the two classifiers are based on a common FC layer. $C^s$ and $C^t$ contain respectively $K$ outputs corresponding to the numbers of categories on the source and the target domains. The baseline has three Softmax classifiers: $C^s$, $C^t$ and $C^{st}$. $C^{st}$ concatenates the inputs from the source of the target domain, i.e., $v^s$ and $v^t$ to a form $[v^s, v^t]\in\mathbf{R}^{2K}$. The classifier $C^{st}$ can discriminate the domain via probability vector $P^{st}\in[0,1]^{2K}$. The SymNets train the classifiers $C^s$ and $C^t$ via cross-entropy loss with the source samples and corresponding labels. The classifier $C^{st}$ is trained with both the source and target samples and domain labels (i.e., $D = [0,1]^{2k}$) to formulate domain probabilities $P^{st}$.

\subsection{Regional Feature Representations.} 

{To form a fine-grained representation of the image features, we split the image features via the channels, akin to the channel-wise attention \cite{chen2017sca}. We then formulate a set of fine-grained feature representations from the grid and channel multi-head groups. specifically, if we split the channel into $H$ heads, and each group has $N$ number of grids, then the total number of the fined-grained features is $T$, described as follows: 
\begin{equation}
\begin{split}
    F & = {ResNet\_50}({Input}),  \\
    I_g &  = \{I_0, ..., I_h, ... I_H\} = Mulithead(F), \\
    I_h & =\{I_{0}, ..., I_{N}\}, \\
    I & = \{I_0, ..., I_T\} = Flatten(I_g),
\end{split}\label{eq:region}
\end{equation}
where $Input$ is the input image, and $F$ is the last convolution feature from the ResNet-50 network.}

\subsection{The Uncertainty Measurement.}
The pseudo-labels generated in a naive way contain much noise, i.e., incorrect classification results, which still need to be handled properly in previous research. Many previous approaches use a confidence-based uncertainty measurement as a choosing standard for pseudo-labels~\cite{rizve2021defense}. In our paper, the pseudo-labels generated in the target domain involve a Triplet loss training for cross-domain matching. As a result, we propose a novel prototype similarity-based uncertainty measurement method. Specifically, we formulate a set of prototypes in the source domain, where each category has one prototype. We obtain the prototype for each category by averaging all the image features from that category, as described,
\begin{equation}
\begin{split}
&   proto_k^s = \frac{\sum_{i=1}^{N_s^k} V^{s}_k(i)}{N_s^k}, \\
\shortintertext{\textit{Practically, an average running algorithm, expressed as:}}
&   proto_k^s(i) = \beta * proto_k^s (i-1) + (1-
\beta) * V^s_k(i), \\ 
& i = 1, ..., N_s^k, 
\end{split}
\end{equation}
where $proto_k^s$ means the prototype for the category $k$ in the source domain $s$ and $N_s^k$ is the number of samples in the source dataset whose category label is $k$. $i$ is the ith iteration in one epoch. $\beta$ is the control coefficient for the running average. Note that we set all the prototypes to zeros when a new training epoch comes and compute the new prototypes in the new epoch. Hence, we obtain a set of prototypes, expressed as $P = \{ proto_k^s | k = 1,..., K\}$, where $K$ is the number of categories in the source domain.

We then calculate the similarity between the prototype set $P$ and the target domain's image features. First, we obtain the temporary classification results of the target samples via classifier $C^t$, which is expressed as
\begin{equation}
\hat{y}^t = {argmax}(C^t(V^t)).
\end{equation}

Subsequently, we compute the cosine similarity between the prototypes and the target features which correspond to the category of each prototype. 

\begin{equation}
{s}_k = {Cosine}({proto}^s_k, V^t[\hat{y}^t_k]), 
\end{equation}
The similarity vector ${s}_k$ is considered the certainty value of the classification results of the classifier in the target domain, i.e., the certainty value of the pseudo-label $\hat{y}^t_k$.

\subsection{Trainable Topk Scheme}
The certainty $C_i, i\in[1;B]$ and the corresponding uncertainty $U_i$ are defined:
\begin{equation}
\begin{split}
& C_i = \left \{ \begin{split} & {s}_i, \ {if} \ {s}_{i} \ in \ {Topk}({s}_{i}), \\
& 0,  {otherwise} \end{split} ,  \right \} \\ 
& U_i = 1 - C_i , i \in [1; B]\\
\end{split}
\end{equation}

{To avoid laborious tuning of the $k$ hyper-parameter in a conventional Topk scheme, we propose a trainable Topk scheme, which is a generic algorithm and should be easily extended to many applications. Specifically, the adaptive Topk are implemented via a Gumbel Softmax and a masking technique to achieve the ability of back-propagation:
\begin{equation}
\begin{split}
& k = Gumbel\_Softmax({s}_i), \\
& {mask} = {Ones}(k-1)  \oplus One\_hot(k) ,\\
& C_i = {mask} \odot {s}_i, i \in [1; B], 
\end{split}
\end{equation}
where Ones$(Dim)$ indicates an all-ones vector with a dimension of $Dim$, $One\_hot$ means the one hot embedding, $\oplus$ is the vector adding operation, and $\odot$ is the element-wise product. With the Gumbel Softmax, the $k$ value is automatically generated and integrated with the training of the whole model; with the masking operation, we make the $k$ value in the Topk scheme trainable, as the operations involved are continuous. }


\subsection{The Adaptive Triplet Loss Learning.}
Though target samples are aligned to the source samples, some target samples might still be near the decision boundary, as illustrated in Figure~\ref{fig:illustration}. This misalignment often occurs between two similar categories, and it is not easy to correct the misclassified
target samples caused by misalignment. Based on the SymNets baseline model and to avoid aligned samples falling into other categories, we introduce an adaptive Triplet loss for cross-domain matching.

As explained previously, similar semantic samples from the source and the target domains should be aligned regardless of the domain difference. However, the lack of available labels prevents the target domain from direct matching. To solve this issue, we apply the uncertainty $U_k$ for each target sample that is classified to the category $k$ (pseudo-label $k$). The higher the uncertainty, the larger margin should be introduced to make the discrimination clearer. The cross-domain matching loss objective is described more formally:
\begin{equation}
\begin{split}
& \mathcal{L}_{Triplet_{st}}(V^s, y^s, V^t, \hat{y}^t) \\
&  = \frac{1}{B} \sum^B_{i=1} U_{i} \odot [\max_{y^s_i=\hat{y}^t_j}||V^s_i - V^t_j||^2 \\ 
& - \min_{y^s_i\neq \hat{y}^t_k}||V^s_i - V^t_k||^2  
 + (\beta + U_{i})]_{+} \\
& + \frac{1}{B} \sum^B_{i=1} U_{i} \odot [\max_{\hat{y}^t_i=y^s_j}||V^t_i - V^s_j||^2 
\\ & - \min_{\hat{y}^t_i\neq y^s_k}||V^t_i - V^s_k||^2 + (\beta + U_{i})]_{+},
\end{split}
\end{equation}
where $V^s$, $V^t$ are the feature from the source and the target domain, respectively. The cross-domain Triplet loss only performs on the pseudo-labels with good certainty and an adaptive margin.

\subsection{The Reinforced Attention Mechanism.}
As shown in Figure~\ref{fig:system}, we first model the attention weights generation process as a finite Markov Decision Process (MDP) and sample a discrete action using Multinomial Sampling. We pre-define $n$ action categories, i.e., $A=\{a_1, a_2, ..., a_n\}$, The state space contains the input region features and the attention weights generated so far, which are $s_t = \{ I^0, Att^{0}..., I^{t-1}, Att^{t-1}\}$. The policy is parametrised via a GRU model to explore the environment and sample the action. More formally:
\begin{equation}
\begin{split}
   & h = GRU(I^t, h^{t-1}), \ \  t= 1,..., T \\
   & a = Softmax(h^t\odot W_{\mu}^t), \\
     &  a_{Sample}^t = Multinomial(a), \\
   & logprob_{a}^t = \log (a[a_{Sample}^t]),  \\
\end{split}\label{mu}
\end{equation}
where $I^i$ is the $i_{th}$ region feature in the $I$, corresponding to Equation~\ref{eq:region}. $GRU$ is the Gated Recurrent Unit (GRU) used to model the attention weights generation problem as MDP. $W_{\mu}^i \in \mathcal{R}_{s \times n}$ are the weights that need to be learned. $Sample$ is the size of the feature vector. 

After we obtain the attention weights, we perform element-wisely multiplication between the hidden features and the attention weights, expressed as follows:
\begin{equation}
\begin{split}
& {Attention_t} = \frac{exp(a_{Sample}^t)}{\sum_t^T exp( a_{Sample}^t)}, \\
& E = \sum_t^T (h_t \odot {Attention_t}),
\end{split}
\end{equation}
where $Attention_t, \ t\in[1;T]$ is the normalized attention weights, and $E$ is the final image embedding.

To be simple and efficient, we formulate the PG as an online learning method, specifically, the REINFORCE algorithm~\cite{williams1992simple}. The PG for the action space is then to maximise the long-term reward with the following expression:
\begin{equation}
\begin{split}
&    \nabla_\theta J(\theta) =  \\
& \mathbb{E}_{\tau \sim \pi_\theta(\tau)}
\left[ \left(\sum_{t=0}^{T} \nabla_\theta \log{\pi_\theta}(a_t \mid s_t)\right) \left(\sum_{t=0}^{T} r(s_t, a_t)\right)\right].
\end{split}\vspace{0.2cm}\label{pgmu}
\end{equation}

We use the one sample Monte-Carlo to approximate the accumulative reward, i.e., $\sum_{t=0}^{T} r(s_t, a_t) = \sum_{t=0}^{T} \mathcal{R}$, where $\mathcal{R}$ is the reward and will be defined later. Also, $\log{\pi_\theta}(a_t \mid s_t) = logprob_{a}^t$, which is obtained from Equation~\ref{mu}. Hence, Equation~\ref{pgmu} can lead to a PG loss function as follows:
\begin{equation}
\begin{split}
&    \mathcal{L}_{PG} = - \sum_{i=1}^{B}
\left[ \left(\sum_{t=0}^{T} \nabla_\theta logprob_{a}^t \right) \left(\sum_{t=0}^{T} \mathcal{R}_i \right)\right],
\end{split}\label{loss_discrete}
\end{equation}
where the reward function $\mathcal{R}$ is defined as follows:
\begin{equation}
\mathcal{R}_i = C_{t} \odot (AP_i (V_i^s, V_k^t) + AP_i(V_i^t, V_k^s)).
\end{equation}

\subsection{Adversarial Domain Alignment}
We include the domain adversarial loss~\cite{ganin2015unsupervised} to align the source and target domain, which is expressed as $\mathcal{L}_{adv}$:
\begin{equation}
\begin{split}
  & DC = {Grad\_Reverse}(Dis(V)),
  \\ & V = F+E, \\
  & \mathcal{L}_{adv} = BCE(DC, Domain\_label),
\end{split}
\end{equation}
where the final embedding $V$ is an addition fusion with the original feature $F$ and $E$, and $Grad\_Reverse$ (as shown in Figure~\ref{fig:system}) is a gradient reversal layer to make the feature in-discriminating on the domain differences. $BCE$ is the binary cross-entropy loss for binary classification.

\subsection{The Overall Losses.}
The overall loss function contains several parts:
\begin{equation}
\begin{split}
    Loss = \mathcal{L}_{SymNets} + \alpha*( \mathcal{L}_{PG} + \mathcal{L}_{Triplet_{st}} + \mathcal{L}_{adv}),
\end{split}
\end{equation}
where $\mathcal{L}_{SymNets}$ is the SymNets baseline's optimization loss functions, $\mathcal{L}_{PG}$ is the policy gradient loss for reinforced attention, $\mathcal{L}_{Triplet_{st}}$ is the inter-domain adaptive Triplet loss, and $\alpha$ controls the contribution of the proposed methods.

\section{Experiments}
In this section, we first introduce the datasets used, followed by implementation details, then the numerical results, and last we present the qualitative evaluation.

\subsection{Datasets}
We perform our experimental evaluation and report results on a mix of standard unsupervised domain adaptation benchmark datasets. 
\subsubsection{Office-31} Office-31~\cite{saenko2010adapting} is a widely-applied dataset for real-world unsupervised domain adaptation. It contains 4,110 images, spanning 31 categories in three domains: Amazon (A), Webcam (W) and DSLR (D). 
\subsubsection{Office-Home} Office-Home is an image classification benchmark dataset~\cite{venkateswara2017deep}, which contains categories of objects found in office and home environments, with 4 domains: The real world (Rw), Clipart (Cl), Product (Pr), and Art (Ar). We report the performance of our model on this dataset. In addition, we perform ablation studies of our method and compare it with State-of-the-arts methods on four domain adaptation tasks. 
\subsubsection{DomainNet} DomainNet is a large unsupervised domain adaptation benchmark, containing 0.6 million images belonging to 6 domains, with 345 categories. Due to labelling noise presenting in its full version, we instead use the subset proposed in Tan et al.~\cite{tan2020class}, which applies 40-commonly seen classes for four domains: Real ({R}), Clipart ({C}), Painting ({P}) and Sketch ({S}). 
\subsubsection{VisDa-2017} The VisDa-2017~\cite{peng2017visda} dataset is the largest synthetic-to-real object classification dataset with over 280k images in the training, validation and testing splits. All three splits share the
same 12 object categories. The training domain consists of 152k synthetic images, which are generated by rendering 3D models of the same object categories from different angles and under different lighting conditions. The validation domain includes 55k images by cropping an object in real images
from COCO~\cite{lin2014microsoft}. The testing domain contains 72k images cropped from video frames in YT-BB~\cite{real2017youtube}.

\begin{table*}[!t]\caption{The results on the Office-31 dataset.}
\centering
\renewcommand\arraystretch{1.25}{
	\resizebox{\linewidth}{!}{
	\begin{tabular}[htb]{llllllll}
		\toprule
		Methods & A$\rightarrow$ W & D$\rightarrow$ W & W$\rightarrow$ D & A$\rightarrow$ D & D$\rightarrow$ A & W$\rightarrow$ A & Avg \\
		\midrule
		ResNet-50~\cite{he2016deep}& 68.4±0.2 &96.7±0.1 &99.3±0.1& 68.9±0.2 & 62.5±0.3 &60.7±0.3 &76.1 \\
		DANN~\cite{ganin2015unsupervised} &82.0±0.4 &96.9±0.2 &99.1±0.1& 79.7±0.4& 68.2±0.4& 67.4±0.5 &82.2 \\
		ADDA~\cite{tzeng2017adversarial} &86.2±0.5 &96.2±0.3 &98.4±0.3& 77.8±0.3 &69.5±0.4 &68.9±0.5 &82.9 \\
		JAN-A~\cite{long2017deep} & 86.0±0.4 &96.7±0.3 &99.7±0.1& 85.1±0.4 &69.2±0.3& 70.7±0.5 &84.6 \\
		MADA~\cite{pei2018multi} &90.0±0.1 &97.4±0.1 &99.6±0.1& 87.8±0.2 &70.3±0.3 &66.4±0.3 &85.2 \\
         Kang et al.~\cite{kang2018deep} & 86.8±0.2 &99.3±0.1 &100.0±.0 &88.8±0.4 &74.3±0.2& 73.9±0.2 &87.2 \\
         CDAN+E~\cite{long2017conditional} &94.1±0.1 &98.6±0.1 &100.0±.0 &92.9±0.2 &71.0±0.3& 69.3±0.3 & 87.7 \\
         SymNets~\cite{zhang2020unsupervised}
          &90.8±0.1 &98.8±0.3 &100.0±.0 & 93.9±0.5 &{74.6±0.6}& 72.5±0.5 &88.4 \\
         \textbf{\textit{AdaTriplet-RA}} (Ours) & \textbf{93.0±0.3} & \textbf{99.2±0.3} & \textbf{100.0±.0} & \textbf{95.2±0.4}  &  \textbf{75.0±0.2} & \textbf{74.1±0.5} & \textbf{89.4}\\
		\bottomrule
	\end{tabular}\label{office_31}
	}}
\end{table*}

\begin{table*}[!t]\Huge \caption{The results on the Office-Home dataset.}
\centering
	\resizebox{\linewidth}{!}{
	\begin{tabular}[htb]{lllllllllllllll}
		\toprule
		Methods & Ar$\rightarrow$ CI & Ar$\rightarrow$ Pr & Ar$\rightarrow$ Rw & CI$\rightarrow$ Ar & CI$\rightarrow$ Pr & CI$\rightarrow$ Rw & Pr$\rightarrow$ Ar & Pr$\rightarrow$ CI & Pr$\rightarrow$ Rw & Rw$\rightarrow$ Ar & Rw$\rightarrow$ CI & Rw$\rightarrow$ Pr & Avg \\
		\midrule
		ResNet-50~\cite{he2016deep}& 34.9& 50.0& 58.0 &37.4 &41.9& 46.2 &38.5 &31.2& 60.4& 53.9 &41.2 &59.9 & 46.1 \\
		DAN~\cite{long2015learning} &43.6 &57.0 &67.9 &45.8 & 56.5 &60.4 &44.0 &43.6 &67.7& 63.1& 51.5& 74.3& 56.3 \\
		DANN~\cite{ganin2015unsupervised} & 45.6 &59.3 &70.1 &47.0 &58.5& 60.9& 46.1& 43.7 &68.5 &63.2 &51.8 &76.8 &57.6 \\
		CDAN+E~\cite{long2017conditional} & \textbf{50.7} &70.6 &76.0 &57.6 &70.0 &70.0 &57.4& 50.9 &77.3 &70.9 & \textbf{56.7} &81.6 &65.8 \\
		SymNets~\cite{zhang2020unsupervised} & 47.7 & 72.9& 78.5& 64.2 & 71.3 &74.2 &{64.2}& {48.8}&79.5& {74.5} &52.6& 82.7& 67.6 \\
		\textbf{\textit{AdaTriplet-RA}} (Ours) &  {49.3} &	\textbf{75.8} &	\textbf{80.4}&		\textbf{67.3}&	\textbf{73.7}&	\textbf{75.8}	&	\textbf{65.6} &	\textbf{50.1}&	\textbf{81.0} &		\textbf{75.3}&	{54.3}&	\textbf{83.1}&		\textbf{69.3} \\
		\bottomrule
	\end{tabular}}\label{office_home} 
\end{table*}

\begin{table*}[!t]\Huge \caption{The results on the DomainNet dataset.}
\centering
	\resizebox{\linewidth}{!}{
	\begin{tabular}[htb]{lllllllllllllll}
		\toprule
		Methods & R $\rightarrow$ C &R$\rightarrow$ P& R$\rightarrow$ S& C$\rightarrow$ R& C$\rightarrow$ P& C$\rightarrow$ S& P$\rightarrow$R &P$\rightarrow$ C &P$\rightarrow$ S &S$\rightarrow$ R &S$\rightarrow$C &S$\rightarrow$ P &AVG \\
		\midrule
		ResNet-50~\cite{he2016deep} & 65.75 &68.84 &59.15 &77.71 &60.60 &57.87& 84.45 & 62.35 &65.07 &77.10 &63.00 &59.72 &66.80 \\
	     BBSE~\cite{lipton2018detecting}  & 55.38 &63.62 &47.44& 64.58& 42.18& 42.36 &81.55 &49.04 &54.10& 68.54 &48.19 &46.07& 55.25 \\
	     PADA~\cite{cao2018partial} &65.91 &67.13& 58.43& 74.69 &53.09 &52.86 &79.84 &59.33 &57.87& 76.52 &66.97& 61.08 &64.48 \\
	     MCD~\cite{saito2018maximum}  &61.97 &69.33 &56.26 &79.78 &56.61 &53.66 &83.38& 58.31& 60.98& 81.74&56.27 &66.78 &65.42 \\
	     DAN~\cite{long2015learning}  &64.36 &70.65 &58.44 &79.44 &56.78 &60.05 &84.56 &61.62 &62.21& 79.69 &65.01 &62.04& 67.07 \\
	     F-DANN~\cite{wu2019domain}  &66.15 &71.80& 61.53& 81.85& 60.06& 61.22& 84.46& 66.81& 62.84& 81.38 &69.62 &66.50 &69.52 \\
	     UAN~\cite{you2019universal} &71.10 &68.90 &67.10 &83.15 &63.30 &64.66 &83.95 &65.35 &67.06 & 82.22 &70.64 &68.09 &72.05 \\
	     JAN~\cite{long2017deep}  &65.57 &73.58& 67.61 &85.02 &64.96 &67.17 &87.06 &67.92 &66.10& 84.54 &72.77& 67.51& 72.48 \\
	     ETN~\cite{cao2019learning} &69.22& 72.14 &63.63 &86.54& 65.33 &63.34 &85.04 &65.69 &68.78 &84.93 &72.17& 68.99& 73.99 \\
	     BSP~\cite{chen2019transferability} &67.29 &73.47& 69.31 &86.50 &67.52 &70.90 &86.83 &70.33 &68.75 &84.34 &72.40 &71.47 &74.09 \\
	     DANN~\cite{ganin2015unsupervised} &63.37 &73.56 &72.63 &86.47& 65.73 &70.58& 86.94 &73.19 &70.15& 85.73& 75.16& 70.04& 74.46 \\
	     COAL~\cite{tan2020class}  &73.85 &75.37 &70.50 &89.63 &69.98 &71.29 &89.81 &68.01 &70.49& 87.97 &73.21 &70.53 &75.89 \\
	     InstaPBM~\cite{li2020rethinking} &80.10 &75.87 &70.84 &89.67 &70.21 &72.76 &89.60 &74.41 &72.19 &87.00 &79.66 &71.75 &77.84 \\
	     ISFDA~\cite{li2021imbalanced} &\textbf{81.52} &77.29 &\textbf{73.55} &90.09 &75.11 &\textbf{74.78} &89.57 &\textbf{76.70} &\textbf{76.07} &{87.55} &\textbf{79.70} &73.13 &79.58 \\
	     SymNets~\cite{zhang2020unsupervised} &  79.17 & 81.54 & 68.61 & 86.43 & 74.37 & 67.53  & 82.65  & 64.74 & 70.11 & 84.18 & 77.07 & 77.70 & 76.18 \\
	     \textbf{\textit{AdaTriplet-RA}} (Ours)   & 79.69	& \textbf{82.79}	&73.34	&	\textbf{91.14} & 	\textbf{78.45} &	74.28	&	\textbf{89.96} &	74.45 &	74.70		& \textbf{88.22} &	{78.04} &    \textbf{79.74} &		\textbf{80.40}\\
		\bottomrule
	\end{tabular}}\label{DomainNet}
\end{table*}

\begin{table*}[!t]\Huge \caption{The results on the VisDa-2017 Test dataset (ResNet-101).}
\centering
	\resizebox{\linewidth}{!}{
	\begin{tabular}[htb]{llllllllllllll}
		\toprule
		Methods & plane &bcycl &bus &car& horse &knife &mcycl &person& plant& sktbrd &train & truck& Avg \\
		\midrule
		ResNet-101~\cite{he2016deep} & 67.7 &36.6 &48.4 &68.2& 76.9& 5.3& 65.8 &38.0& 72.5 &29.1 &82.1 &3.73 &49.5 \\
		DANN~\cite{ganin2015unsupervised} & 87.1 &63.0& 76.5& 42.0 &90.3 &42.9 &85.9 &53.1 &49.7 &36.3 &85.8 &20.7 &61.1 \\
        DAN~\cite{long2015learning} & 81.9 &77.7 &82.8 &44.3 &81.2 &29.5 &65.1 &28.6 &51.9 &54.6 &82.8 &7.8 &57.4 \\
        JAN-A~\cite{long2017deep} &75.7& 18.7 &82.3 &86.3 &70.2 &56.9 &80.5 &53.8 &92.5 &32.2 &84.5 &54.5 & 65.7 \\
        MCD~\cite{saito2018maximum} &87.0 &60.9 &83.7 &64.0 &88.9 &79.6 &84.7 &76.9 &88.6 &40.3 &83.0 &25.8 &71.9 \\
        ADR~\cite{saito2018adversarial} &87.8 &79.5 &83.7 &65.3 &92.3 &61.8 &88.9 &73.2 &87.8 &60.0 &85.5 &32.3 &74.8 \\
        BSP~\cite{chen2019transferability} &92.4 &61.0 &81.0 &57.5 &89.0 &80.6 &90.1 &77.0 &84.2 &77.9 &82.1 &38.4 &75.9 \\
        SWD~\cite{lee2019sliced} &90.8 &82.5 &81.7 &70.5 &91.7 &69.5 &86.3 &77.5 &87.4 &63.6 &85.6 &29.2 &76.4 \\
        DADA~\cite{tang2020discriminative} &92.9 &74.2 &82.5 &65.0 &90.9 &93.8 &87.2 &74.2 &89.9 &71.5 &86.5 &48.7 &79.8 \\
        IterLNL~\cite{zhang2021unsupervised} &89.0 &79.5 &84.3 &81.0 &87.7 &88.1 &\textbf{92.5} &38.7 &87.1 &\textbf{96.9} &78.8 &67.0 &80.9 \\
        STAR~\cite{lu2020stochastic} &95.0 &84.0 &84.6 &73.0 &91.6 &91.8 &85.9 &78.4 &94.4 &84.7 &87.0 &42.2 &82.7 \\
        SE~\cite{french2018self} &\textbf{95.9} &\textbf{87.4} &85.2& 58.6& \textbf{96.2}& \textbf{95.7} &90.6& 80.0 &\textbf{94.8} & 90.8& 88.4& 47.9 & 84.3 \\
        SymNets~\cite{zhang2020unsupervised} & 89.8 &	39.1	&82.8	&92.7	&79.0	&18.3	&81.9 &	\textbf{89.1}	&91.8	&23.7	&91.6	&76.2	&75.6 \\
        \textbf{\textit{AdaTriplet-RA}} (Ours) &92.3&	62.9&	\textbf{87.2} &	\textbf{94.5}	&85.6	&73.1 &	88.2 &	83.4&	93.8&	83.6	&\textbf{93.0}&	\textbf{77.0}&	\textbf{85.3}\\
		\bottomrule
	\end{tabular}}\label{visda2017}
\end{table*}

\begin{table}[!t]\Huge \caption{The results on the VisDa-2017 Test dataset (ResNet-50).}
\centering
	\resizebox{0.35\linewidth}{!}{
	\begin{tabular}[htb]{llllllllllllll}
		\toprule
		Methods & Avg \\
		\midrule
		ResNet-50~\cite{he2016deep} &40.2 \\
        DAT~\cite{ganin2016domain} &63.7 \\
        GTA~\cite{sankaranarayanan2018generate} &69.5 \\
        MCD~\cite{saito2018maximum} &69.2 \\
        CDAN~\cite{long2018conditional} &70.0 \\
        DEC~\cite{zhu2021source} &73.3 \\
        CAMCD~\cite{azzam2021unsupervised} &73.6 \\
        SymNets~\cite{zhang2020unsupervised} &70.8 \\
        \textbf{\textit{AdaTriplet-RA}} (Ours) &	\textbf{77.0}\\
		\bottomrule
	\end{tabular}
 }
 \label{visda2017_50}
\end{table}


\begin{table*}[!t]\caption{Ablation study on the Office-Home dataset.}
\centering
	 \resizebox{\linewidth}{!}{
\renewcommand\arraystretch{1.25}
	\begin{tabular}[htb]{llllllll}
		\toprule
		Methods & Ar$\rightarrow$ CI & Ar$\rightarrow$ Pr & Ar$\rightarrow$ Rw & CI$\rightarrow$ Ar & Avg\\
		\midrule
		Baseline (SymNets) & 45.48 & 72.70 & 78.29 & 63.91  & 65.10\\
		Ours + Adv. &  46.87 & 74.13 & 78.74 & 66.13 & 66.47 \\
		Ours + Adv. + RA & 47.90 & 73.37 & 79.10 & 66.34 & 66.68 \\
	
		Ours + Adv. + RA + Triplet ($k$=20) & 47.12 & 74.75 & 79.02 & 66.70 & 66.90 \\
		Ours + Adv. + RA + Triplet ($k$=1)  & 46.53 & 74.43 & 78.79 & 65.43 & 66.30 \\
		Ours + Adv. + RA + Triplet ($k$=10) & 47.33 & 74.66 & 79.27 & 65.76 & 66.80 \\
        \hline
		Ours w/ $\alpha$=1 & 46.53 & 74.29 & 78.54 & 65.76 & 66.28 \\
        Ours w/ $\alpha$=5 & 47.63 & 74.79 & 79.00 & 66.87 & 67.07 \\
		Ours w/ $\alpha$=10 & 48.18 & 75.11 & 79.18 & 67.04 & 67.38 \\
		\hline
        Ours w/ H=1 & 48.18 & 75.11 & 79.18 & 67.04 & 67.38 \\
        Ours w/ H=2 & 48.43 & 75.13 & 79.92 & 67.04 & 67.63\\
        Ours w/ H=4 & 48.50 & 75.44 & 80.10 & 67.28 & 67.83\\
        Ours w/ H=8 & 49.26 & 75.76 & 80.35 & 67.28 & 68.16\\
        Ours w/ H=16 & 48.01 & 75.40 & 79.57 & 67.12 & 67.53\\
		\hline
		Ours + Adv. + RA + Triplet (Adp. $k$) & \textbf{49.26} & \textbf{75.76} & \textbf{80.35} & \textbf{67.28} & \textbf{68.16} \\
		\bottomrule
	\end{tabular}\label{ablation}
}
\end{table*}

\begin{figure*}
\centering
{
\includegraphics[width=0.85\linewidth, height=4cm]{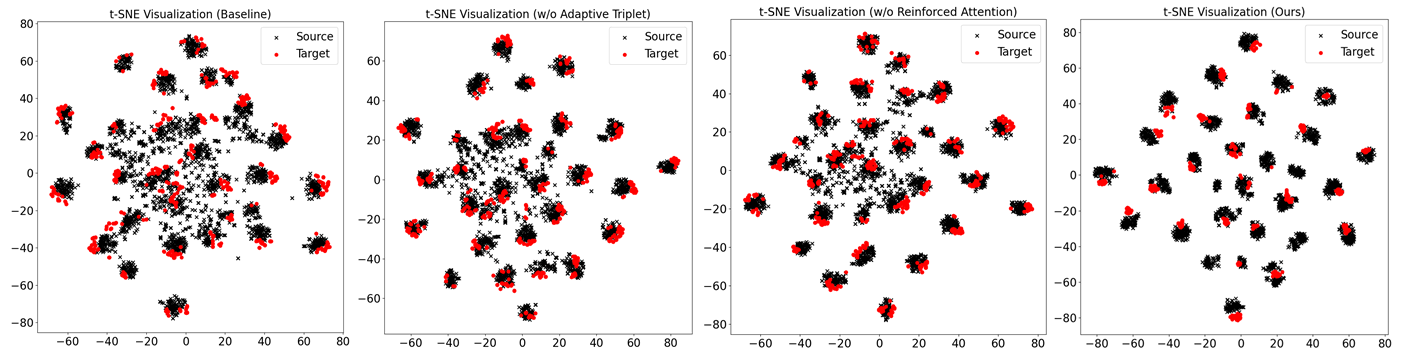}  
}     
\caption{t-SNE Visualisation of different methods on Office-31 dataset A$\rightarrow$W task, our method has a better clustering quality than other approaches in ablation studies.}  
\label{fig:vis}     
\end{figure*}

\begin{figure}   
    \centering
    \includegraphics[width= \linewidth]{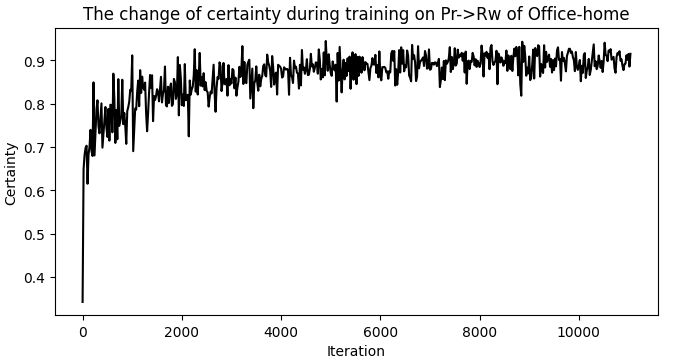}
    \caption{Visualisation of the certainty values in the training process on the Office-Home dataset, in which we can see the certainty increases as the training is performed, indicating more reliable pseudo-labels will be selected. }
    \label{fig:certainty}
\end{figure}

\begin{figure}
    \centering
    \includegraphics[width= 0.8\linewidth]{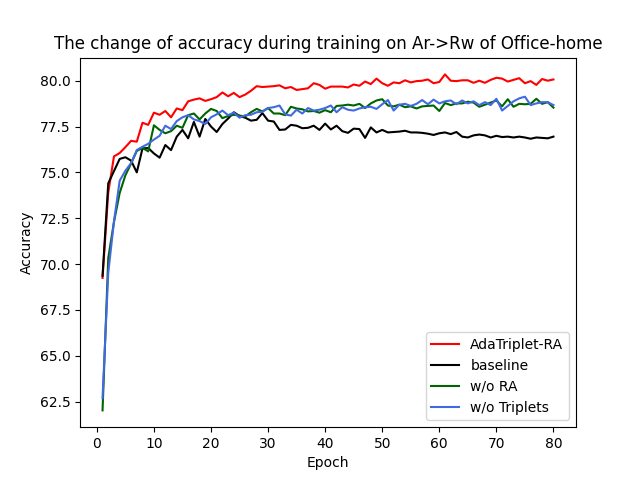}
    \caption{{Visualization of the convergence situation on the Office-Home dataset. Our method has a better convergence speed and higher results than the baseline. We also observe that the combination of "Triplet" and "RA" can boost the final convergence speed and accuracy results. }}
    \label{fig:ab_accuracy}
\end{figure}

\begin{figure*}
    \centering
    \includegraphics[width=\linewidth]{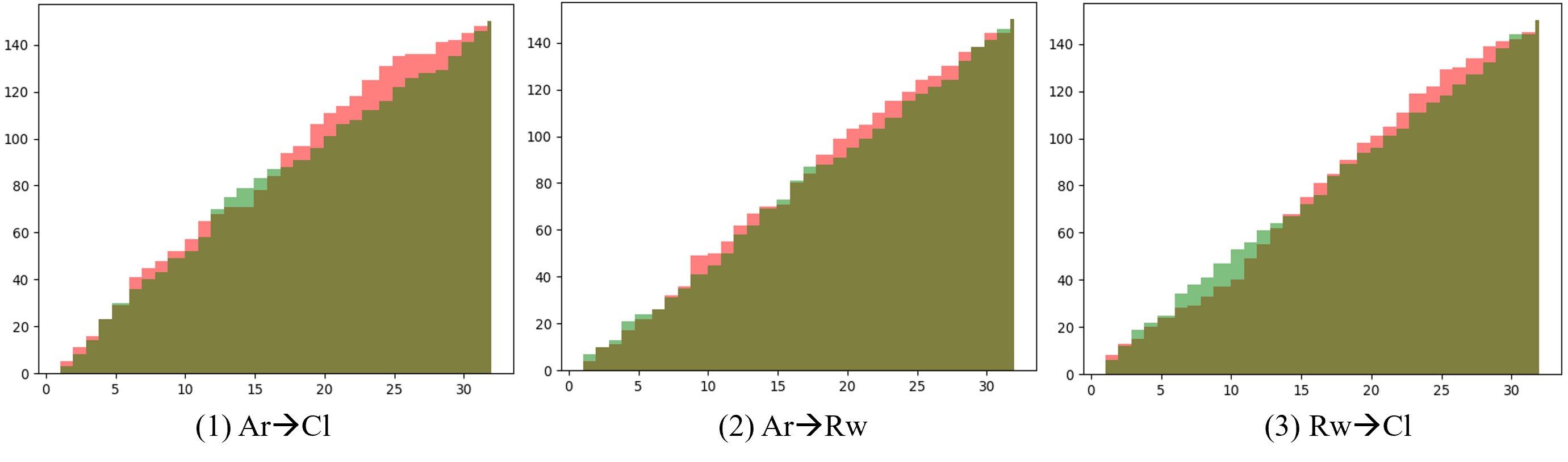}
    \caption{Accumulated histogram visualization for the $k$ values in the Office-Home dataset. The green colour indicates the $k$ distribution of the initial 150 iterations of the training, while the red corresponds to the last 150 iterations of the training stage. The deep green colour is the overlap of the two distributions. We observe that the red colour has a bigger coverage in the distribution, indicating that the later stage selects more reliable pseudo-labels, given the comparison of the $k$ distributions. 
    }
    \label{fig:topk}
\end{figure*}

\begin{figure*}
    \centering
    \includegraphics[width=\linewidth]{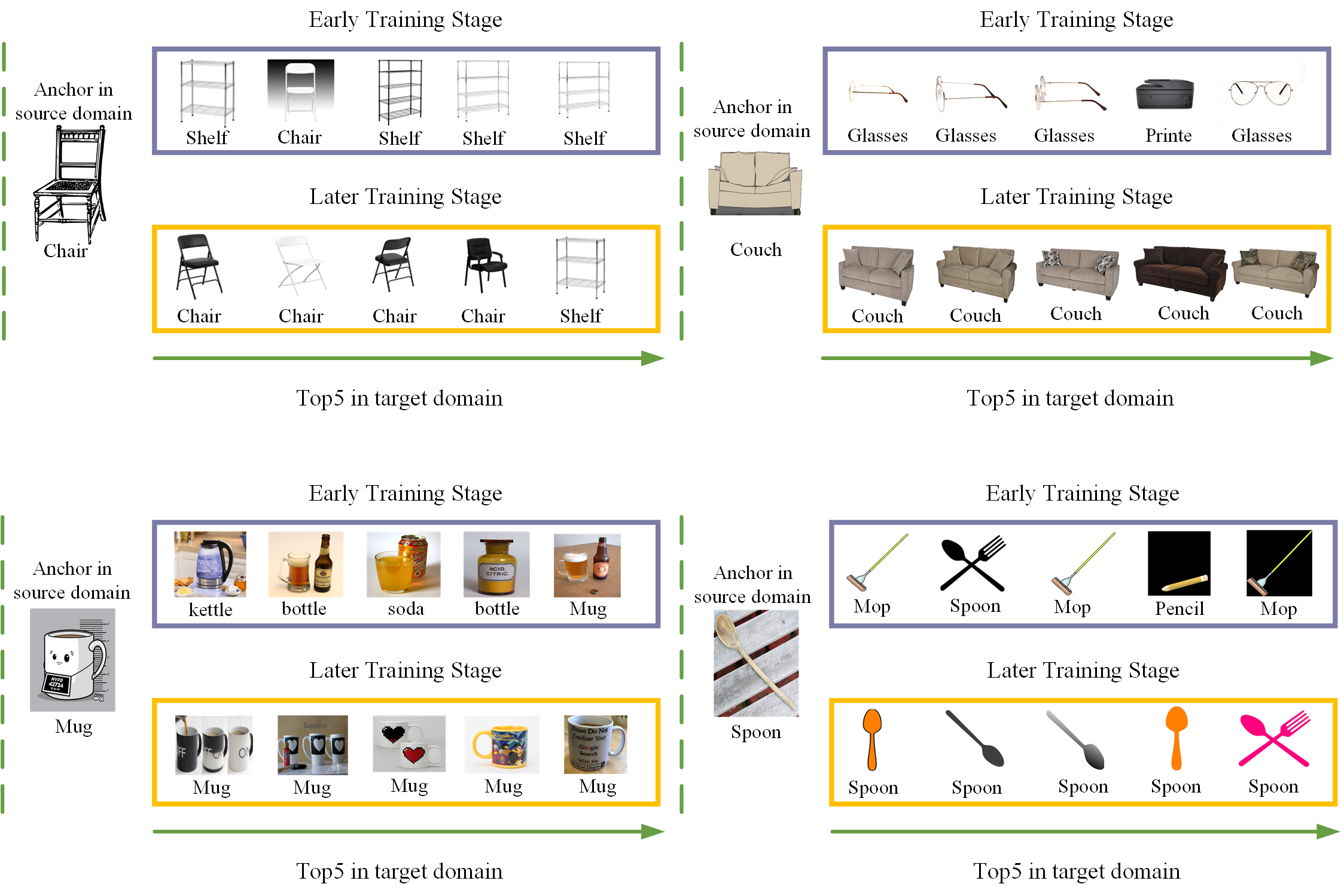}
    \caption{Visualisation of some examples for the inter-domain sample matching effect. Based on the cosine similarity metric, we select the top-5 samples from the target domain given the source anchor image. In the early training stage, the anchor in the source domain hardly finds a match in the target domain, while the sample matching is significantly improved in the later training stage. This phenomenon indicates that the training for sample matching is successful. }
    \label{fig:matching}
\end{figure*}

\subsection{Implementation Details}
\paragraph{Baseline model}
We build our model based on the ResNet-50 backbone Network~\cite{he2016deep}. We apply the SymNets training method as our baseline model~\cite{zhang2019domain}. All the training losses from the initial SymNets are kept in our training, but our model's network and training configurations are different. 
\paragraph{Configurations}
Our model can be trained in a single Nvidia Geforce 2080-TI with an 11GB memory GPU card. The configurations are as follows: 
\begin{itemize}

\item  For both the source and target domain, the batch size is 32. In the Triplet loss and reinforced attention, we set the hyper-parameter $\beta$ as 0.5.

\item The dimension of the input and hidden features of the GRU and the embedding dimension of the adaptive Triplet loss are all 2048. The dimension of the input, hidden features of the GRU, and the embedding dimension of the adaptive Triplet loss are also 2048.

\item We use the SGD optimiser, the early stopping technique, to select the best model checkpoint. All the SGD optimisers are with Nesterov momentum and weight decay of 5e-4. On all the datasets, we apply a learning rate of 1e-3 to train the model with 80 epochs. 

\item On both the large-scale DomainNet and VisDa-2017 datasets, we have a certain number of fixed layers during training (the first convolutional block in the ResNet). 

\end{itemize}

\paragraph{Environment}
Our model is built on PyTorch-1.10 platform~\cite{paszke2017automatic}. We conducted all our experiments on a PC equipped with Nvidia Geforce 2080-TI GPU and installed Windows-10 and a CUDA, Cudnn, from the Nividia toolkit.

\subsection{Comparison with State-of-the-art Methods}
We compare our methods with other State-of-the-art approaches on all three datasets. The comparison of the Office-31 dataset is shown in Table~\ref{office_31}. Our average accuracy surpasses all the previous results. Specifically, we lead the SymNets~\cite{zhang2020unsupervised} method by 1.0\% in average precision of all six domain transfer tasks. Note that our model is just as efficient as SymNets though we introduce extra modules. Even with a limited computing resource and a smaller batch size (32) (compared with 128 in SymNets implementations), we improve the results of current State-of-the-art methods. The comparison of Office-Home is presented in Table~\ref{office_home}, where a similar phenomenon is observed. Our model's average accuracy in the Office-Home dataset leads to State-of-the-art by 1.7\%. The subsequent ablation studies also prove the improving impact of each ingredient of the proposed model.

The results of DomainNet are illustrated in Table~\ref{DomainNet}. 
Our method outperforms SymNets~\cite{zhang2020unsupervised} in all twelve domain transition tasks and increases SymNets' average accuracy by 4.22\%.
 
Interestingly, the distribution of our accuracy results regarding the different domain adaptation tasks is quite different from the existing State-of-the-art method, i.e., ISFDA~\cite{li2021imbalanced}, which has a similar intuition of selective optimisation. ISFDA~\cite{li2021imbalanced} mainly performs the selection based on class balance whilst we propose a novel way of certainty measurement. Our method tends to perform better or close in most domain adaptation tasks, with 80.40\% average accuracy.

We report the results of our methods on the VisDa-2017 test set in Table~\ref{visda2017} and Table~\ref{visda2017_50}. In addition, we report the results on the ResNet-50 and ResNet-101 backbone networks. Our methods improve the SymNets baseline, by a large margin, with both backbone networks. Especially our methods based on ResNet-101 won the SE~\cite{french2018self}, which is the champion in the VisDa-2017 challenge, significantly. Interestingly, our methods with the ResNet-50 backbone surpass most methods with the ResNet-101 backbone network (including the SymNets baseline), which validates the superiority of the proposed sample matching scheme.

\subsection{Comparative Study on Each Block of the Model}
\subsubsection{Baseline} As shown in Table~\ref{ablation}, the baseline model which utilizes the same optimization techniques with SymNets~\cite{zhang2020unsupervised} yields poor accuracy results, even worse than the original results~\cite{zhang2020unsupervised}. Our implementation has a smaller batch size (32 versus 128 in~\cite{zhang2020unsupervised}).

\subsubsection{Adversarial training.} From Table~\ref{ablation}, the ``Ours + Adv." is the model with adversarial training. When adding the adversarial domain confusion, the accuracy increases, which validates the effectiveness of the adversarial domain confusion.

\subsubsection{Reinforced attention} From Table~\ref{ablation}, the ``Ours + Adv. + RA." is the model with reinforced attention. When adding the reinforced attention, the accuracy increases, which validates the effectiveness of the attention mechanism.

\subsubsection{Triplet Loss} As shown in Table~\ref{ablation}, the scheme ``Ours + Adv. + RA + Triplet ($k$=20)" has an obvious positive impact over the baseline, which validates the effectiveness of the domain matching scheme.

\subsubsection{Fixed value of $k$} Both the accuracy of a single task and the average accuracy are improved as $k$ increasing. $k$ is a critical parameter in this model, as it controls the pseudo labels certainty threshold, impacting the Triplet loss performance and reinforced attention. Although with great significance, $k$ is not extremely sensitive, with a small performance alternation on different values. Still, note that a tiny value of $k$ quickly loses functional pseudo-labels and thus deteriorates the model to a scheme similar to "Ours + Adv. + RA". 

\subsubsection{Adaptive $k$}{To avoid the manual tuning of the $k$ hyper-parameter, the adaptive $k$ is with critical significance. One does not need to hand-tune the $k$ parameter, which expands the application scope and efficiency of the model; In addition, the adaptive $k$ can bring a significant performance gain, as shown in "Ours + Adv. + RA + Triplet (Adp. $k$)", yielding the best results in the ablation study.}

\subsubsection{ The value of $H$}
{We report the comparative study of the number $H$, i.e., the channel splitting factor. A suitable $H$ is good for finding fine-grained features which benefit the overall performance. 
From Table \ref{ablation}, we observe that the $H$ is not a very sensitive hyper-parameter for the final performance. 8 for $H$ yields the best results.}

\subsubsection{Coefficient $\alpha$} The coefficient $\alpha$ is critical in maintaining good performance. A reasonably larger $\alpha$ increases the proportion of adaptive Triplet loss and reinforced attention in the model's training, which helped produce a better performance of adaptive Triplet loss and reinforced attention.



\subsection{Visualization}
\subsubsection{Cluster visualization}
We perform t-SNE Visualisation for different methods on Office-31 "Amazon-to-Webcam" (A$\rightarrow$W) task, illustrated in Figure~\ref{fig:vis}. On Office-31 "Amazon-to-Webcam" (A$\rightarrow$W) task, there are 2817 data in the Amazon domain as source data and 795 data in the Webcam domain as target data. Ours can have a better visualisation result than the baseline and other methods in ablation studies. Especially the scheme "w/o Adaptive Triplet" has better visualisation results than the baseline. The scheme "w/o Reinforced Attention" is better in cluster visualisation. Our full model has the best visualisation quality, which matches the numerical results.

\subsubsection{Certainty and $k$ visualization}
To see the change of certainty value and $k$ during the training stage and validate our adaptive scheme, we visualise the certainty value during training as shown in Figure \ref{fig:certainty} and the distribution of $k$ in Figure \ref{fig:topk}. The certainty value gradually increases as the training is performed, which shows that the model tends to become more confident on the pseudo-labels in the target domain. We apply a cumulative histogram to visualise the distribution of the $k$ values. We compare the $k$ value's distribution between the initial 150 iterations and the last 150 iterations of the training process. As shown in Figure \ref{fig:topk}, the model tends to select more large value $k$ in the later stage of the training (the red colour covers more area than the green colour), indicating the increasing robustness.

{\subsubsection{Convergence speed visualization}
To visualise the convergence speed, we plot the training accuracy in each epoch, as shown in Figure~\ref{fig:ab_accuracy}. As the training epoch increases, our model tends to have a higher convergence speed than the baseline. The positive impact of the Triplet training and the reinforced attention upon convergence is also validated.}

\subsubsection{Matching performance visualization}
To see the change in the inter-domain sample matching performance during the training, we visualise the matched sample by selecting the Top-5 similar examples via cosine similarity, as shown in Figure \ref{fig:matching}. We compare the matching performance between the early stage of the training and the trained model. It is clear from the figure that the trained model tends to select more correct samples from the target domain, which illustrates that our method improves the inter-domain sample matching performance.

\section{Conclusions}
    This paper improves the unsupervised domain adaptation through a novel perspective, i.e., improving the sample-level discriminating capability. To this end, we propose an uncertainty-aware inter-domain sample matching scheme. We utilise an uncertainty-aware adaptive Triplet loss and reinforced attention to fulfil the domain match. This novel perspective and the corresponding technical solutions effectively improve the unsupervised domain adaptation task. {In addition, the proposed modules, such as the trainable adaptive Topk module, the adaptive Triplet loss, and the Reinforced attention, are all model-agnostic, which can easily be plugged and applied in many other applications.} Comprehensive experiments validate the effectiveness of the uncertainty-aware domain match, with State-of-the-art results achieved on several publicly available benchmark datasets.

\section{Acknowledgments}
This work is supported by the National Natural Science Foundation of China (62106110, U20B2061), the Natural Science Foundation of Jiangsu (BK20210646) and the Research Innovation Program for College Graduates of Jiangsu (KYCX22\_1209). 

\bibliographystyle{elsarticle-num}
\bibliography{sp}
\end{document}